\documentclass[10pt,twocolumn,letterpaper]{article}

\usepackage[pagenumbers]{wacv} 

\usepackage{times}
\usepackage{epsfig}
\usepackage{graphicx}
\usepackage{amsmath}
\usepackage{bbm}
\usepackage{amssymb}
\usepackage{multirow, tikz, makecell, verbatim, booktabs, array}
\usepackage{cuted}
\usepackage[hypcap=false]{caption}
\usepackage[labelsep=period]{caption}

\usepackage[pagebackref,breaklinks,colorlinks]{hyperref}

\newcommand\Tstrut{\rule{0pt}{2.6ex}}         
\newcommand\Bstrut{\rule[-0.9ex]{0pt}{0pt}}   
\usepackage{multirow, tikz, makecell, verbatim, booktabs, array}
\newcommand\norm[1]{\lVert#1\rVert}

\usepackage[capitalize]{cleveref}
\crefname{section}{Sec.}{Secs.}
\Crefname{section}{Section}{Sections}
\Crefname{table}{Table}{Tables}
\crefname{table}{Tab.}{Tabs.}

\def\wacvPaperID{272} 
\def\confName{WACV}
\def\confYear{2024}

\begin{document}

\title{Point-DynRF: Point-based Dynamic Radiance Fields from a Monocular Video}


\author{Byeongjun Park \qquad Changick Kim \\
Korea Advanced Institute of Science and Technology (KAIST)\\
{\tt\small \{pbj3810, changick\}@kaist.ac.kr}
}
\maketitle

\begin{strip}
\centering
\includegraphics[width=0.3\textwidth]{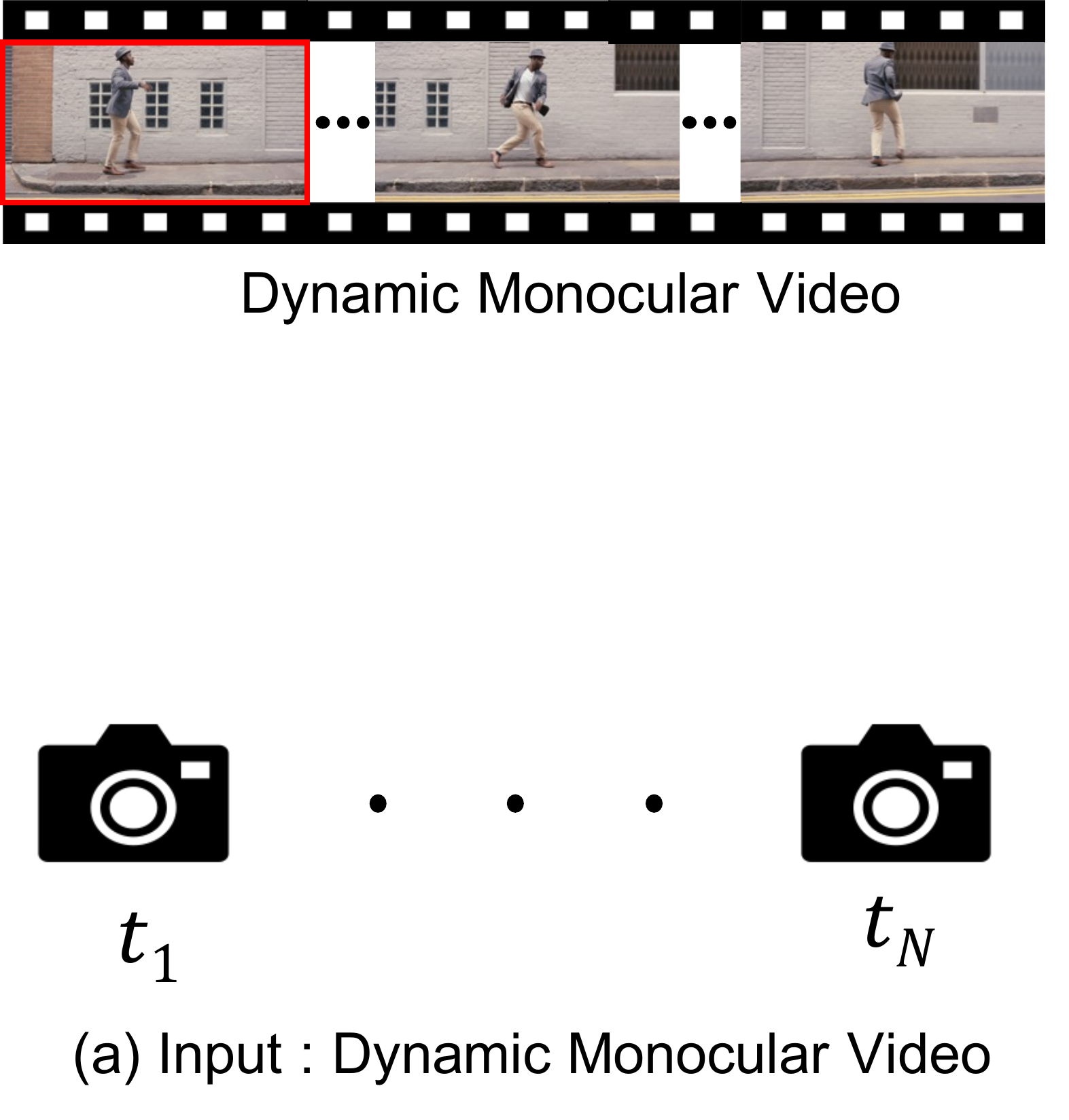} \hspace{0.2cm}
\begin{tikzpicture}
\draw (0, 0) node[inner sep=0]{\includegraphics[width=0.6\linewidth]{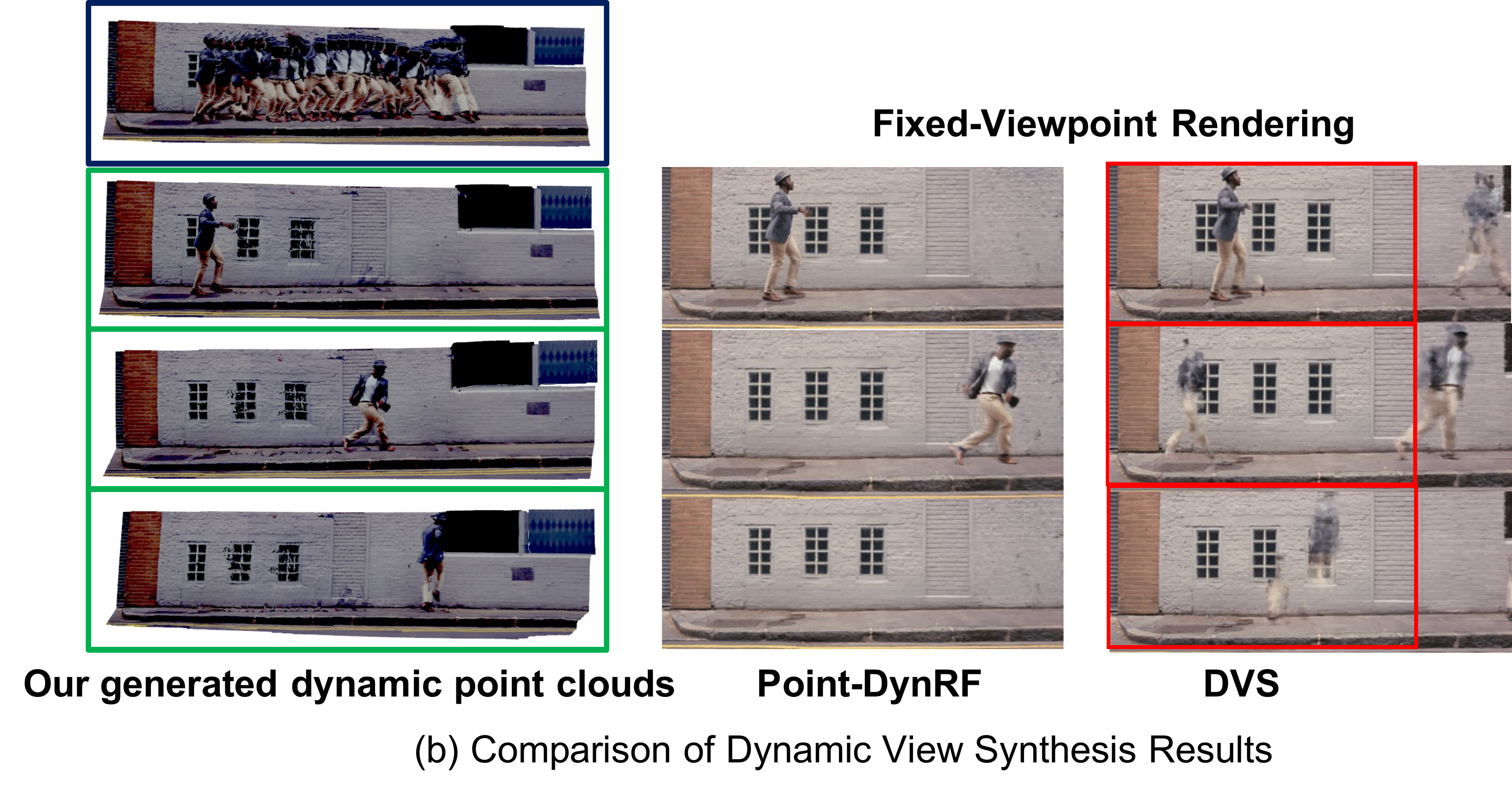}};
\draw ( 4, -2) node {\small {\cite{gao2021dynamic}}};
\end{tikzpicture}
\vspace{-0.3cm}
\captionof{figure}{\textbf{Point-based Dynamic Radiance Fields for Long-Term Novel View Synthesis.} Point-DynRF takes a monocular video following dynamic objects, as shown in (a), and uses neural 3D points generated from the input video to efficiently represent dynamic radiance fields. (b) We design a novel framework that samples a subset point cloud (\textcolor{green}{green boxes}) at each time step from the entire point cloud (\textcolor{blue}{a blue box}) and regresses dynamic radiance fields only on the scene surface where the subset point cloud are located. Especially with a wide-range camera trajectory, Point-DynRF addresses the duplicating problem of the state-of-the-art method (\textcolor{red}{red boxes}). \label{fig_summary}}
\vspace{-0.2cm}
\end{strip}

\begin{abstract}
    Dynamic radiance fields have emerged as a promising approach for generating novel views from a monocular video. 
    However, previous methods enforce the geometric consistency to dynamic radiance fields only between adjacent input frames, making it difficult to represent the global scene geometry and degenerates at the viewpoint that is spatio-temporally distant from the input camera trajectory.
    To solve this problem, we introduce point-based dynamic radiance fields (\textbf{Point-DynRF}), a novel framework where the global geometric information and the volume rendering process are trained by neural point clouds and dynamic radiance fields, respectively.
    Specifically, we reconstruct neural point clouds directly from geometric proxies and optimize both radiance fields and the geometric proxies using our proposed losses, allowing them to complement each other.
    We validate the effectiveness of our method with experiments on the NVIDIA Dynamic Scenes Dataset and several causally captured monocular video clips.
\end{abstract}
\section{Introduction}
\label{sec:intro}
Consider a monocular video recording of dynamic objects.
While it is challenging to distinguish between static and dynamic areas in a single frame, analyzing the entire video sequence enables us to differentiate the background from the moving objects.
Moreover, we can also predict the background outside a captured frame by assuming that the background scene remains constant over time.
This scene reasoning ability enables us to identify the moving objects and integrate partially available scene information, which is crucial for understanding in-the-wild videos and scaling the free-viewpoint rendering.

Existing novel view synthesis methods for monocular videos often use separate modules for static and dynamic regions, where view-dependent radiance fields are designed for static regions and time-dependent radiance fields for dynamic regions~\cite{yoon2020novel, li2021neural, gao2021dynamic, park2021hypernerf, wu2022d, tretschk2021non, liu2023robust, li2023dynibar}.
In this regard, recent deformable NeRFs~\cite{park2021hypernerf, wu2022d, TiNeuVox, pumarola2021d} learn sufficient view dependencies from small camera trajectories to represent the background, while representing the remaining regions using time-dependent radiance fields.
However, in the real world, there are many cases where the camera does not follow a narrow trajectory, and deformable NeRFs fail to distinguish between the background and dynamic objects due to the lack of learning view dependencies.

On the other hand, flow-based methods~\cite{li2021neural, gao2021dynamic, liu2023robust, li2023dynibar} use additional supervisions from pre-trained depth~\cite{ranftl2021vision}, optical flow~\cite{teed2020raft} and semantic segmentation~\cite{he2017mask} estimation networks to constrain the radiance field since identifying moving objects and estimating their motion in monocular videos are challenging.
By imposing geometric constraints on the radiance field, flow-based methods can design dynamic radiance fields for large scenes.
Despite its scalability, we observe that flow-based methods quickly degenerate for viewpoints in spatio-temporally distant from the input camera trajectory, and the generated image is blurry and sometimes contain duplicated objects.
This is because time-dependent radiance fields are trained by the optical flow supervision to satisfy geometric consistency between adjacent frames, which fails to incorporate global geometric information of entire scene from wide-range camera trajectories.
Figure~\ref{fig_summary}-(b) shows the problem of a state-of-the-art dynamic view synthesis method~\cite{gao2021dynamic} where a person is duplicated outside of the input frame and the background is not preserved after the person walks by because of the duplicated person.

Motivated by our observations, we introduce point-based dynamic radiance fields (\textbf{Point-DynRF}) to represent the entire scene geometry and produce more realistic long-term novel view synthesis results.
Point-DynRF is built upon the Point-NeRF~\cite{xu2022point} representation, which reconstructs 3D neural point clouds and encodes the localized scene representation from neighboring neural points.
While Point-NeRF aims at static scenes, we extend it to consider the time domain where different subsets of neural point clouds are sampled at each time step to represent time-varying radiance fields.
Specifically, we utilize a pre-trained depth estimation network~\cite{ranftl2021vision} and pre-defined foreground masks~\cite{gao2021dynamic} to initialize pixel-wise depth and rigidness of our neural point clouds, respectively.
Moreover, we propose a dynamic ray marching, where we march a ray over a subset of the entire point cloud consisting of all background points and the dynamic points corresponding to the rendering time.
As each subset of neural point clouds represents the actual scene surface of the corresponding rendering time, our Point-DynRF can regress dynamic radiance fields only on the scene surface at that rendering time and alleviate to generate of duplicated dynamic objects.

To train Point-DynRF, we simply modify the training objective of DVS~\cite{gao2021dynamic} and jointly optimize the neural point clouds and dynamic radiance fields, rather than solely supervising the radiance fields using initialized depth and foreground masks.
Specifically, we train Point-DynRF to align the initialized learnable depth and foreground masks with the volume rendered depth and dynamicsness maps.
Through the joint optimization scheme, the global scene geometry and dynamic radiance fields are further refined and complement each other, addressing the degeneration problems of previous methods in long-term dynamic view synthesis.
Extensive experiments on the NVIDIA Dynamic Scenes~\cite{yoon2020novel} and several monocular video clips show the efficiency and effectiveness of our method.
\begin{figure*}[t!]
  \centering
  \includegraphics[width=0.98\linewidth]{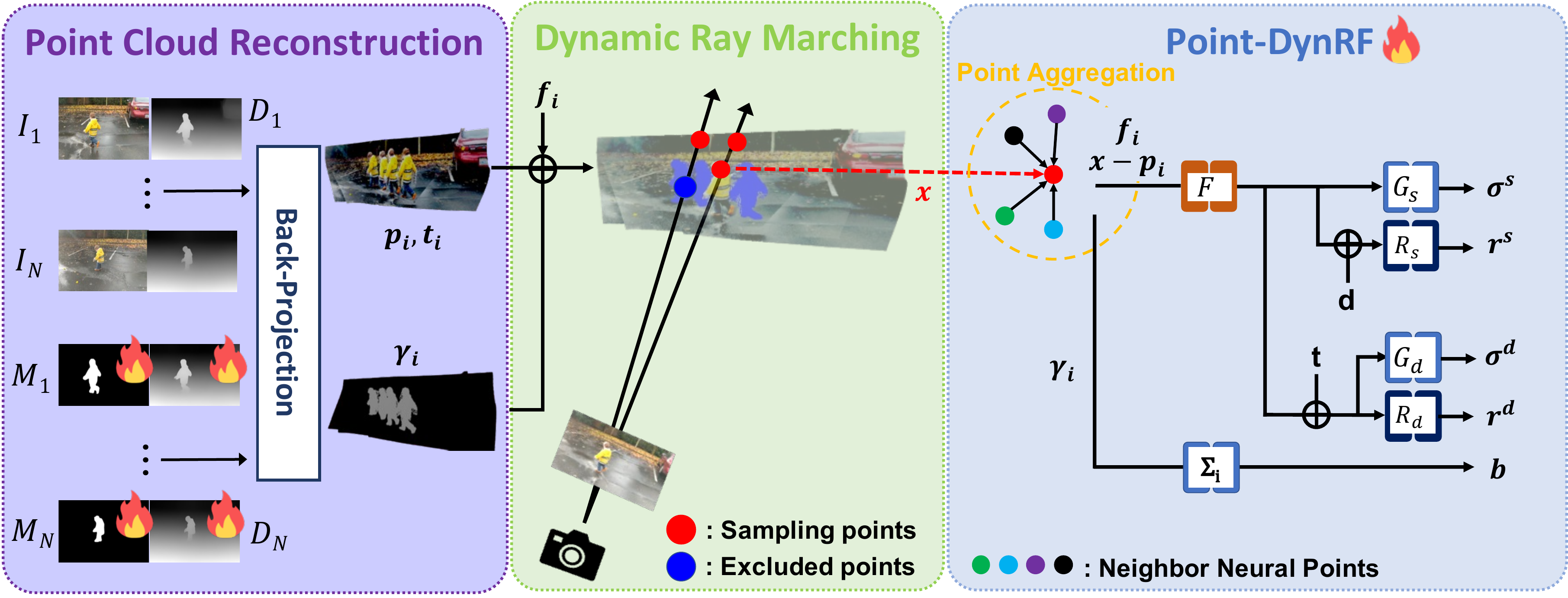}
  \vspace{-0.2cm}
  \caption{
  \textbf{An overview of network architecture.} Our framework consists of three components. First, we initialize per-frame depth maps $D_{n}$ and foreground masks $M_{n}$ for a given $N$ frames. Then, we back-project each pixel of $N$ frames to reconstruct our neural 3D point clouds. Each neural point $i$ contains its spatio-temporal locations $(\textbf{p}_{i},t_{i})$, a point-wise rigidness $\gamma_{i}$, a randomly initialized neural feature vector $f_{i}$ to represent the local scene representation. Then, we select a subset point cloud at a rendering time step $t$ and assign sampling points where the ray meets the neural points as they march. Finally, we regress a volume density and a radiance on both view-dependent and time-dependent radiance fields. The volume density and radiance for each sampling point in the ray are integrated via volume rendering to output an RGB color.
  }
  \label{fig_network}
  \vspace{-0.6cm}
\end{figure*}

\section{Related Works}
\label{sec:related}

\paragraph{Neural representations for novel view synthesis.}
Novel view synthesis aims to generate new views of a scene given multiple posed images.
To consider the arbitrary viewpoints in three-dimension, multiple-view geometry is often utilized and combined with image-based rendering methods to synthesize realistic novel views~\cite{debevec1996modeling,levoy1996light,seitz2006comparison,zitnick2004high,debevec1998efficient}.
Moreover, deep neural networks have been explored to improve the visual quality of novel views by using explicit geometric proxies, such as multi-plane image~\cite{srinivasan2019pushing, zhou2018stereo,wizadwongsa2021nex}, point cloud~\cite{aliev2020neural, wu2020multi, wang2020neural}, and voxel~\cite{sitzmann2019deepvoxels, TiNeuVox}.

Recently, coordinate-based neural representations~\cite{chen2019learning, mescheder2019occupancy, mildenhall2021nerf, park2019deepsdf} have achieved outstanding results in modeling the scene as implicit scene representations.
In the context of novel view synthesis, Neural Radiance Fields (NeRF)~\cite{mildenhall2021nerf} has been proposed to model the scene as a continuous volumetric field with neural networks.
The success of NeRF is attributed to the extension of neural representation design, which facilitates free-viewpoint rendering with various applications, such as relighting~\cite{boss2021nerd}, appearance editing~\cite{liu2021editing, yuan2022nerf}, reflections~\cite{guo2022nerfren}, and generative models~\cite{chan2021pi, niemeyer2021giraffe, cai2022pix2nerf}.
Despite its remarkable scalability, several methods~\cite{xu2022point, kurz2022adanerf} focus on the fact that NeRF samples a large number of unnecessary points for each ray.
Specifically, Point-NeRF~\cite{xu2022point} models a volumetric radiance field with 3D neural point clouds, avoiding ray sampling in the empty space and encoding localized scene representations.
Our work extends Point-NeRF, encoding different scene representations for static and dynamic regions by leveraging its capability to encode localized scene representations. 

\paragraph{Dynamic view synthesis for videos.}
Dynamic view synthesis focuses on generating novel views with dynamically moving objects at arbitrary viewpoints and time stamps.
Several works have been proposed to model time-varying scenes on multiple time-synchronized videos~\cite{bansal20204d, stich2008view, zitnick2004high, li2022neural}, sparse camera views~\cite{Huang_2018_ECCV, du2021neural}, stereo camera~\cite{attal2020matryodshka}, and specific domain~\cite{carranza2003free, habermann2019livecap, weng2022humannerf}.
However, modeling neural scene representation from a monocular video is more challenging since it contains a single viewpoint for each time stamp.
This causes ambiguities that radiance can be changed in either a view-dependent or a time-varying or both.
To solve this ambiguity, Yoon~\etal~\cite{yoon2020novel} combines an explicit depth estimation module to leverage geometric transformations (i.e., warping) and to blend strategies for synthesizing novel views of a dynamic scene, but it requires a time-consuming preprocessing to generate manually annotated foreground masks.
Recently, flow-based methods~\cite{gao2021dynamic, li2021neural, xian2021space, liu2023robust} directly regress 4D space-time radiance fields by using additional geometric proxies, such as depth~\cite{ranftl2021vision} and optical flow~\cite{teed2020raft} estimation networks.
Geometric proxies are used as additional supervision to learn their deformation module and constrain temporal changes of a dynamic scene.
Several methods~\cite{wu2022d, park2021hypernerf, tretschk2021non, pumarola2021d, park2021nerfies, TiNeuVox} propose deformable neural radiance fields by modeling a canonical template radiance field and a deformation field for each frame.
Our work also uses geometric proxies for point cloud initialization, but we optimize the dynamic radiance fields and geometric proxies together based on the volume rendering process.
Moreover, point-based dynamic radiance fields allow us to incorporate the entire scene geometry and regress the radiance fields from the actual scene surface for each rendering time.
\section{Method}
Given a monocular video $V = \{I_{1}, I_{2}, \dots, I_{N}\}$ consisting of $N$ frames, our goal is to synthesize novel views at arbitrary viewpoints and time steps. 
To achieve this, we design point-based dynamic radiance fields as shown in Fig.~\ref{fig_network}.
Our model is built on the Point-NeRF~\cite{xu2022point} representation and extends it to consider time-varying radiance fields.
We briefly describe the volume rendering formulation in \ref{method:formul} and then explain how to extend Point-NeRF to consider the time domain in~\ref{method:Point-DynRF}.
Finally, we illustrate the optimization scheme of Point-DynRF in~\ref{method:optimization}.

\subsection{Volume rendering \label{method:formul}}

We construct continuous volumetric fields for modeling dynamic scenes, following the formulation in NeRF~\cite{mildenhall2021nerf}.
Given the camera center $\textbf{o} \in \mathbb{R}^3$ and viewing direction $\textbf{d} \in \mathbb{R}^2$, each pixel's RGB color $\textbf{C} \in \mathbb{R}^3$ is computed by marching a ray $\textbf{r}(s) = \textbf{o} + s\textbf{d}$ through the pixel and approximate the integration over radiance and its volume density $\{(r_{j}, \sigma_{j}) \in \mathbb{R}^3 \times \mathbb{R} \mid j=1, \dots, M\}$ for $M$ sampling points in the ray as:

\begin{equation}
    \textbf{C}(\textbf{r}) = \sum_{j=1}^{M}{T_{j}(\alpha(\sigma_{j}\delta_{j}))r_{j}},
    \label{eq:volume_rendering}
\end{equation}
\begin{equation}
    T_{j} = \text{exp}(-\sum_{k=1}^{j-1}{\sigma_{k}\delta_{k}}),
\end{equation}
where $\alpha(x) = 1 - \text{exp}(-x)$ outputs the opacity at each sampling point, $\delta_{j}$ is the distance between two adjacent sampling points and $T_{j}$ represents a volume transmittance.

\subsection{Point-DynRF Representation \label{method:Point-DynRF}}

Point-NeRF~\cite{xu2022point} is pre-trained on a multi-view stereo dataset~\cite{jensen2014large} or uses only points located on the actual surface with high confidence from COLMAP.
In dynamic scenes, however, it fails to accurately regress the scene geometry since dynamic objects disrupt to estimate point-to-point correspondences.
To solve this ambiguity, we propose Point-DynRF with associated neural point clouds, which are initialized by imprecise depth maps and pre-defined foreground masks, and jointly optimize scene geometry and dynamic radiance fields.

\paragraph{Neural Point Clouds Reconstruction.}
Our neural point clouds are reconstructed by depth maps $\{D_{1}, ..., D_{N}\}$ and foreground masks $\{M_{1}, ..., M_{N}\}$.
We first initialize per-frame depths by using disparity maps $disp_{n}$ obtained from DPT~\cite{ranftl2021vision} and convert it to depth maps with per-frame scale $s_{n}$ and shift $b_{n}$ values as:

\begin{equation}
D_{n}(p) = s_{n} / (disp_{n}(p) + b_{n}).
\end{equation}

Note that we design a more stable network by optimizing scale, shift, and disparity together rather than optimizing pixel-wise depth values individually.
Per-frame foreground masks are obtained as same as DVS~\cite{gao2021dynamic}, and we directly parameterize our point-wise rigidness $\gamma$ with $1$ for the background and $0$ for moving objects.
Thus, we reconstruct neural point clouds as $\mathbb{P} = \{(\textbf{p}_{i}, t_{i}, \textbf{f}_{i}, \gamma_{i}) \mid i =1, ..., L\}$, where each point $i$ is located at $\textbf{p}_{i}$ and captured at time steps $t_{i}$ with a point-wise rigidness $\gamma_{i}$.
We also use a neural feature vector $\textbf{f}_{i}$, which are randomly initialized and parameterized to encode local scene representations.
Since each neural point is a one-to-one match to each pixel of input frames, training the $\textbf{p}_{i}$ and $\gamma_{i}$ of each neural point optimizes the depth and foreground masks.

\paragraph{Dynamic Ray Marching.}
Our dynamic radiance fields are regressed from a different subset of the entire point cloud set $\mathbb{P}$ at each time step based on the sampling time and the point-wise rigidness.
Specifically, we select neural points where their point-wise rigidness is higher than the threshold $\lambda=0.5$, or its temporal location is the same as the sampling time as:
\begin{equation}
\mathbb{P}_{t} = \{(\textbf{p}_{i}, t_{i}, \textbf{f}_{i}, \gamma_{i}) \in \mathbb{P} \mid t_{i} = t \ \text{or} \ \gamma_{i} > \lambda \},
\end{equation}
where neural points with $\gamma_{i}$ is higher than $\lambda$ to be background points to represent the static region whether the position of the dynamic object changes with each subset. 
Moreover, dynamic neural points do not represent the scene surface from different viewpoints, resulting in avoiding unnecessary ray sampling and not duplicating objects.

\paragraph{Neural Point Aggregation.}
After we select the subset of the neural point cloud, Point-DynRF aggregates neural points to output the density and radiance for each shading point.
Specifically, we follow the Point-NeRF~\cite{xu2022point} to query $K=8$ neighbor neural points for ray sampling, and we encode per-point local scene features with an MLP layer $F$ for each shading point \textbf{x} as:

\begin{equation}
f_{i, \textbf{x}} = F(\textbf{f}_{i}, \textbf{x} -\textbf{p}_{i}).
\end{equation}

\paragraph{Volume Density Regression.}
We use density regression MLP layers $G_{s}$ and $G_{d}$ for static and dynamic regions, respectively.
We first encode per-point time-invariant volume density $\sigma^{s}$ and time-varying volume density $\sigma^{d}$ as:

\begin{equation}
\sigma^{s}_{i, \textbf{x}} = G_{s}(f_{i, \textbf{x}}),
\end{equation}
\begin{equation}
\sigma^{d}_{i, \textbf{x}} = G_{d}(f_{i, \textbf{x}}, t).
\end{equation}

Then, the time-invariant volume density $\sigma^{s}_{\textbf{x}}$ and time-variant volume density $\sigma^{d}_{\textbf{x}}$ at the sampling point \textbf{x} is regressed as:
\begin{equation}
\sigma^{s}_{\textbf{x}} = \sum_{i} \sigma^{s}_{i, \textbf{x}}\frac{w_{i}}{\sum w_{i}},
\end{equation}
\begin{equation}
\sigma^{d}_{\textbf{x}} = \sum_{i} \sigma^{d}_{i, \textbf{x}}\frac{w_{i}}{\sum w_{i}},
\end{equation}
where $w_{i} = \frac{1}{\lvert p_{i} - x\rvert}$ is for a distance-based weighted sum that gives higher weight to neural points closer to \textbf{x}.

\paragraph{Radiance Regression.}
We regress a view-dependent radiance $r^{s}_{\textbf{x}}$ and a time-dependent radiance $r^{d}_{\textbf{x}}$ by using MLP layers $R_{s}$ and $R_{d}$, respectively, as:

\begin{equation}
r^{s}_{\textbf{x}} = R_{s}(\sum_{i} \frac{w_{i}}{\sum w_{i}} f_{i, \textbf{x}}, d),
\end{equation}
\begin{equation}
r^{d}_{\textbf{x}} = R_{d}(\sum_{i}\frac{w_{i}}{\sum w_{i}} f_{i, \textbf{x}}, t),
\end{equation}
where $d$ and $t$ is the viewing direction and sampling time, respectively.

\paragraph{Blending Weight Regression.}
We directly regress blending weights from the point-wise rigidness $\gamma_{i}$ of neighboring points as:

\begin{equation}
b_{\textbf{x}} = \mathbbm{1}[\sum_{i}(\frac{w_{i}}{\sum w_{i}} (1 - \gamma_{i})) > \lambda],
\end{equation}
where $\mathbbm{1}$ equals to one if the condition is true.
We define the blending weight as $0$ or $1$ so that either static or dynamic radiance fields dominate at each shading point.
To optimize $\gamma_{i}$, we use the gradient clamping used in Point-NeRF to $\sum_{i}(\frac{w_{i}}{\sum w_{i}} (1 - \gamma_{i}))$ if $\text{MAX}(\sigma^{s}_{\textbf{x}}, \sigma^{d}_{\textbf{x}})$ is larger than a threshold $0.7$ and there exists at least one dynamic point.

\subsection{Training Objectiveness \label{method:optimization}}

In this section, we briefly demonstrate how we jointly optimize dynamic radiance fields and neural 3D point clouds.
Specifically, we introduce reconstruction losses to learn combined NeRF, static NeRF, and dynamic NeRF in Sec.~\ref{sec:loss_recon}, scene geometry losses to reconstruct accurate neural points in Sec.~\ref{sec:scene geometry losses} and joint optimization of Point-DynRF and neural 3D points in Sec.~\ref{sec:joint_optim}.

\subsubsection{Reconstruction Loss}
\label{sec:loss_recon}

\paragraph{Combined NeRF}
We apply a reconstruction loss to dynamic radiance fields, which are a blend of view-dependent and time-dependent radiance fields.
To this end, we combine two radiance fields with blending weights as:

\begin{equation}
    \textbf{C}(\textbf{r}, t, \mathbb{P}_{t}) = \sum_{j=1}^{M}{T_{j}}\left( \alpha(\sigma^{s}_{j}\delta_{j}) (1-b_{j})r^{s}_{j} + \alpha(\sigma^{d}_{j}\delta_{j}) b_{j} r^{d}_{j}\right),
\end{equation}
\begin{equation}
    T_{j} = \text{exp}\left(-\sum_{k=1}^{j-1}{((\sigma^{s}_{k} (1 - b_{k}) + \sigma^{d}_{k} b_{k})\delta_{k}})\right),
\end{equation}
where $\textbf{C}(\textbf{r}, t, \mathbb{P}_{t})$ is a volume rendered RGB value from a ray $\textbf{r}(s) = \textbf{o} + s\textbf{d}$, rendering time $t$, and a subset point cloud $\mathbb{P}_{t}$.
To ensure that the dynamic radiance fields accurately reconstruct the input video sequence, we jointly train view-/time-dependent radiance fields by applying a reconstruction loss $L^{full}_{rec}$ as:

\begin{equation}
L^{full}_{rec} = \sum^{N}_{i=1}\sum_{uv}\norm{\textbf{C}(\textbf{r}^{i}_{uv}, i, \mathbb{P}_{i}) - I^{i}_{u, v}}^{2}_{2},
\end{equation}
where $\textbf{r}^{i}_{uv}$ is a ray for pixel coordinates $(u, v)$ in $i$-th frame and $I^{i}_{u, v}$ is a ground-truth RGB value for pixel coordinates $(u, v)$ in $i$-th frame.

\begin{figure}[t!]
  \centering
  \includegraphics[width=0.97\linewidth]{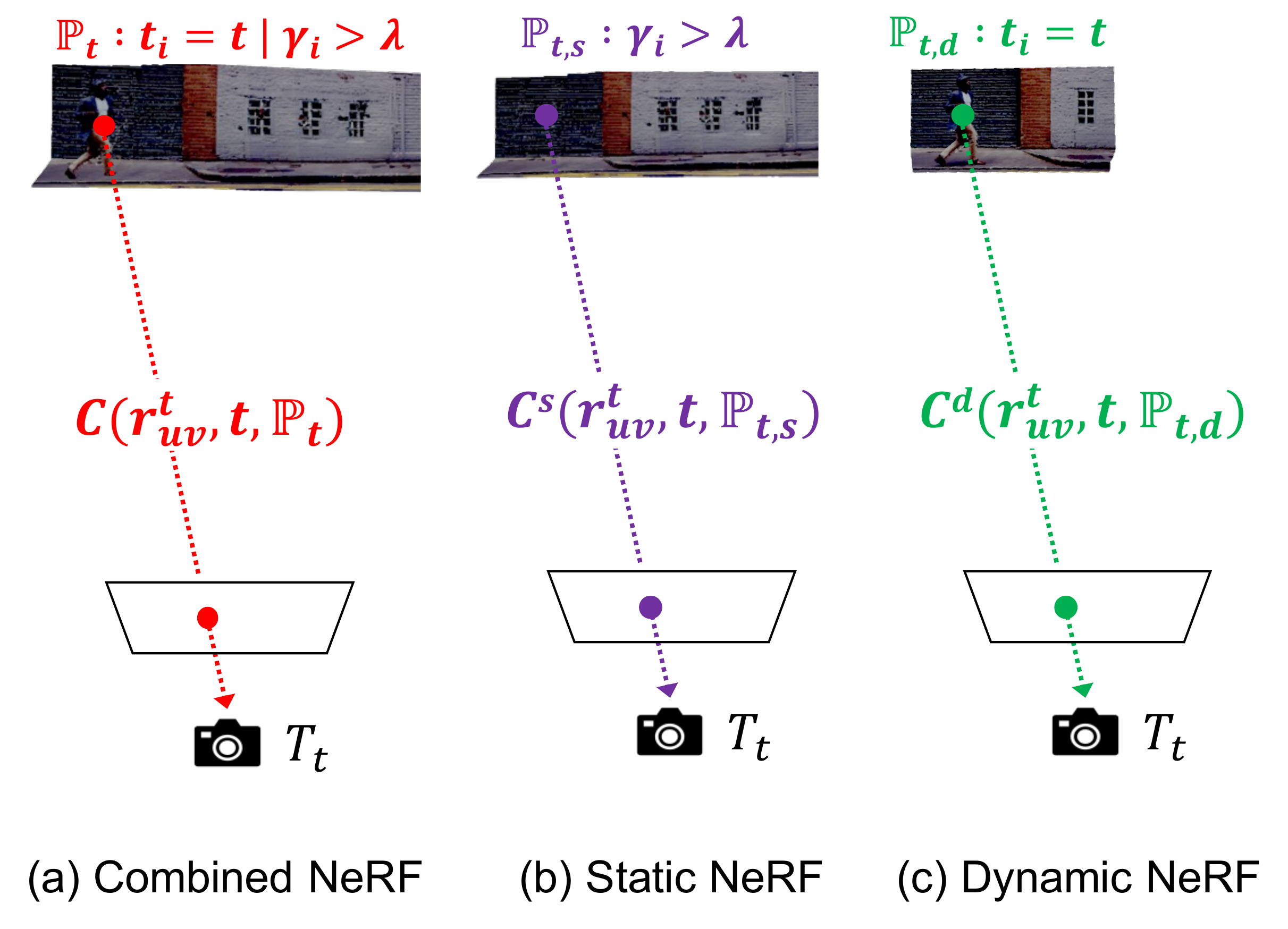}
  \vspace{-0.3cm}
  \caption{
  \textbf{An overview of point cloud subsets for each NeRF.}}
  \label{fig_nerf_type}
\vspace{-0.6cm}
\end{figure}

\paragraph{Static and Dynamic NeRF}
We leverage point-based neural scene representations to learn time-invariant radiance fields (Static NeRF) and time-variant radiance fields (Dynamic NeRF), respectively.
If we sample a subset point cloud $\mathbb{P}_{t, s}$ consisting of only background points as:

\begin{equation}
\mathbb{P}_{t, s} = \{(\textbf{p}_{i}, t_{i}, \textbf{f}_{i}, \gamma_{i}) \in \mathbb{P} \mid \gamma_{i} > \lambda \},
\end{equation}
a volume rendered image contain only the background with no dynamic objects.
Likewise, if we sample a subset point cloud $\mathbb{P}_{t, d}$ captured at a specific time $t$ as:

\begin{equation}
\mathbb{P}_{t, d} = \{(\textbf{p}_{i}, t_{i}, \textbf{f}_{i}, \gamma_{i}) \in \mathbb{P} \mid t_{i} = t \},
\end{equation}
Point-DynRF can render an image restricted to only the neural points at that moment.
Figure~\ref{fig_nerf_type} shows which subset point clouds are used by combined NeRF, Static NeRF, and Dynamic NeRF.
Thus, each radiance field is regressed by using Eq.~\ref{eq:volume_rendering} as:
\begin{equation}
    \textbf{C}^{s}(\textbf{r}, t, \mathbb{P}_{t, s}) = \sum_{j=1}^{M}{T^{s}_{j}(\alpha(\sigma^{s}_{j}\delta_{j}))r^{s}_{j}},
\end{equation}
\begin{equation}
    \textbf{C}^{d}(\textbf{r}, t, \mathbb{P}_{t, d}) = \sum_{j=1}^{M}{T^{d}_{j}(\alpha(\sigma^{d}_{j}\delta_{j}))r^{d}_{j}}.
\end{equation}

Then, we apply reconstruction losses to each radiance field as:

\begin{equation}
L^{s}_{rec} = \sum_{i, u, v}\norm{\textbf{C}^{s}(\textbf{r}^{i}_{uv}, t, \mathbb{P}_{t, s}) - I^{i}_{u, v}}^{2}_{2} * \mathbbm{1}[M^{i}_{u, v} > \lambda],
\end{equation}

\begin{equation}
L^{d}_{rec} = \sum_{i, u, v}\norm{\textbf{C}^{s}(\textbf{r}^{i}_{uv}, t, \mathbb{P}_{t, d}) - I^{i}_{u, v}}^{2}_{2},
\end{equation}
where we only apply $L^{s}_{rec}$ to background regions by using a foreground mask, and $\mathbbm{1}[M^{i}_{u, v} > \lambda]$ indicates whether a rigidness value of pixel coordinates $(u, v)$ in $i$-th frame is higher than the threshold $\lambda$.
Finally, our reconstruction loss is formulated as:
\begin{equation}
L_{rec} = \lambda^{full}_{rec}L^{full}_{rec} + \lambda^{s}_{rec}L^{s}_{rec} + \lambda^{d}_{rec}L^{d}_{rec}.
\end{equation}

\subsubsection{Scene Geometry Loss}
\label{sec:scene geometry losses}

Initialized depth maps well represent the scene geometry but contain scale ambiguities with other frames.
Therefore, we use optical flow maps $f_{gt}$ from RAFT~\cite{teed2020raft} to supervise scale $s_{t}$ and shift $b_{t}$ by applying a flow loss $L_{flow}$ only for background pixels as:

\begin{equation}
[u', v', z']^{T} = T^{-1}_{t'}T_{t}D_{t}[u, v, 1]^{T},
\end{equation}
\begin{equation}
L_{flow} = \sum_{uv}\norm {(\frac{u'}{z'} - u, \frac{v'}{z'} - v) - f_{gt}} * \mathbbm{1}[M^{i}_{u, v} > \lambda],
\end{equation}
where $t'$ indicates a time step for adjacent frames and $T_{t}$ is known camera parameters at $t$.
Note that we detach the gradient from back-propagating to the disparity so that only the scale and shift values can be trained from the flow loss.

\begin{figure}[t]
  \centering
  \includegraphics[width=0.97\linewidth]{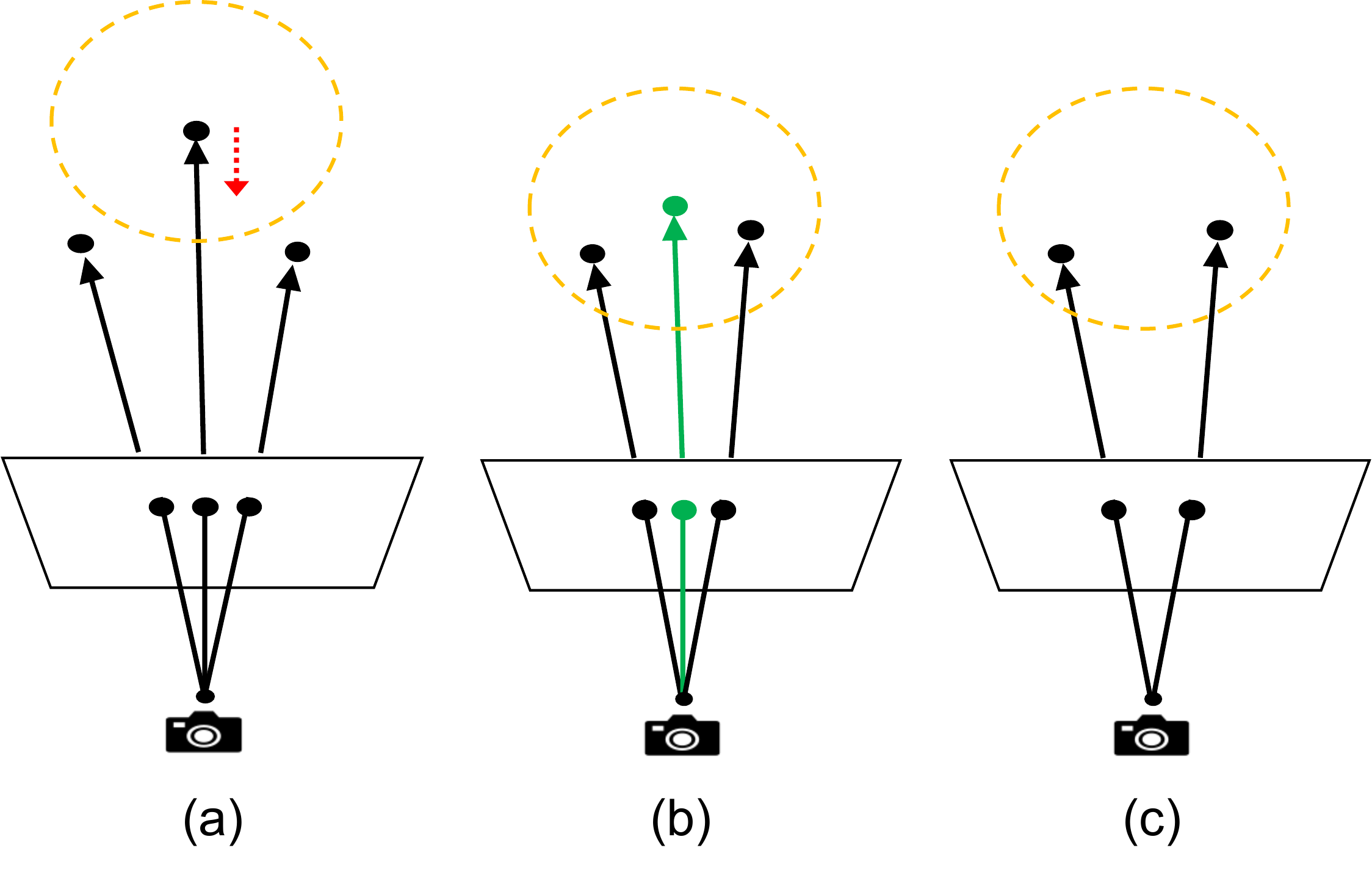}
  \vspace{-0.2cm}
  \caption{\textbf{Missing rays.} Assuming a ray is marched when it has three neighbor points for a shading point. (a) If the depth value is too large, the distance between neighboring pixels in 3D world coordinates is larger than the querying radius and fails to march the ray. Moreover, the subset point cloud is changed as the render time varies, causing rays can be marched (b) or sometimes not (c).}
  \label{fig_miss_ray}
  \vspace{-0.6cm}
\end{figure}

Moreover, we observe two cases that point-based ray marching can miss the ray as shown in Fig.~\ref{fig_miss_ray}.
While learning the scene geometry, some pixels may have large depth values to satisfy the geometric consistency.
As a result, neighboring pixels in the image plane are also outside the query boundary, resulting in the ray can not be marched.
Therefore, we introduce $L^{s}_{miss}$, which is an $\ell_2$-loss to minimize the depth value corresponding to the pixel for which the ray is not marched.
Also, rays may not be marched for different render times in the fixed-viewpoint due the dynamic ray sampling.
To solve this problem, we introduce $L^{d}_{miss}$, which is also an $\ell_2$-loss to maximize the rigidness of a green point in Fig.~\ref{fig_miss_ray}-(b) to be one.
Note that $L^{s}_{miss}$ and $L^{s}_{miss}$ are introduced to deal with outlier cases, since missing rays are rarely present in the entire training process.
Consequently, a scene geometry loss $L_{geo}$ is formulated as:

\begin{equation}
L_{geo} = \lambda_{flow}L_{flow} + \lambda^{s}_{miss}L^{s}_{miss} + \lambda^{d}_{miss}L^{d}_{miss}.
\end{equation}

\subsubsection{Joint Optimization}
\label{sec:joint_optim}
We further introduce loss functions that optimize the dynamic radiance field and neural points together.
Our joint optimization losses are formulated in the same manner as DVS~\cite{gao2021dynamic}.
However, we make a modification by introducing learnable per-frame depth and foreground masks, in contrast to the supervised learning approach of matching the volume-rendered depth $\textbf{\~{D}}(\textbf{r}, t, \mathbb{P}_{t})$ and dynamicsness map $\textbf{\~{M}}(\textbf{r}, t, \mathbb{P}_{t})$ to the initialized depth and foreground mask.

\paragraph{Depth Adjust Loss}


We apply a depth adjust loss $L_{depth}$ to train the depth map of $i$-th frame $D_{i}$ to match the expected depth $\textbf{\~{D}}(\textbf{r}, t, \mathbb{P}_{t})$ as:

\begin{equation}
L_{depth} = \sum^{N}_{i=1}\sum_{uv}\norm{\textbf{\~{D}}(\textbf{r}^{i}_{uv}, i, \mathbb{P}_{i}) - D^{i}_{u, v}}^{2}_{2},
\end{equation}
where $D^{i}_{u, v}$ is a depth value of pixel coordinates $(u, v)$ in $i$-th frame.

\paragraph{Mask Adjust Loss}
Similar to expected depth maps, we use volume rendering for the blending weight to get the dynamicness map $\textbf{\~{M}}(\textbf{r}^{i}_{uv}, i, \mathbb{P}_{i})$ and propose a mask adjust loss $L_{mask}$ to match the per-frame foreground mask.

\begin{equation}
L_{mask} = \sum^{N}_{i=1}\sum_{uv}\norm{\textbf{\~{M}}(\textbf{r}^{i}_{uv}, i, \mathbb{P}_{i}) - M^{i}_{u, v}}^{2}_{2}.
\end{equation}

\begin{table*}[t!]
    \caption{\textbf{Quantitative results on NVIDIA Dynamic Scene dataset~\cite{yoon2020novel}.} Image quality is measured by PSNR and LPIPS. Furthermore, we show the average performance over all view changes at the end. Best results in each metric are in \textbf{\textcolor{red}{bold}}, and second best are  \textcolor{blue}{\underline{underlined}}.}
   \vspace{-0.2cm}
    \centering
    \resizebox{\linewidth}{!}{
        \begin{tabular}{l|cccccc||c}
        \Xhline{3\arrayrulewidth}
        \multirow{2}{*}{Methods}  & \multicolumn{7}{c}{PSNR$(\uparrow)$ / SSIM$(\uparrow)$ / LPIPS$(\downarrow$)}\Tstrut\Bstrut \\ \cline{2-8} 
        & Jumping & Skating & Truck & Balloon1 & Balloon2 & Playground & Avg \Tstrut\Bstrut \\ \hline 
        NeRF~\cite{mildenhall2021nerf} + time  & 16.6 / 0.42 / 0.48 & 19.1 / 0.46 / 0.54 & 17.1 / 0.39 / 0.40  & 17.5 / 0.40 / 0.29 & 19.8 / 0.54 / 0.22 &	13.7 / 0.18 / 0.44	& 17.3 / 0.40 / 0.40\Tstrut \\
        D-NeRF~\cite{pumarola2021d}  & 21.0 / 0.68 / 0.21 & 20.8 / 0.62 / 0.35 & 22.9 / 0.71 / 0.15 & 18.0 / 0.44 / 0.28 & 19.8 / 0.52 / 0.30 & 19.4 / 0.65 / 0.17 &  20.4 / 0.59 / 0.24 \\
        NR-NeRF~\cite{tretschk2021non} & 19.4 / 0.61 / 0.29 & 23.2 / 0.72 / 0.23 & 18.8 / 0.44 / 0.45 & 17.0 / 0.34 / 0.35 & 22.0 / 0.70 / 0.21 & 14.3 / 0.19 / 0.33	& 19.2 / 0.50 / 0.33 \\
        HyperNeRF~\cite{park2021hypernerf} & 17.1 / 0.45 / 0.32 & 20.6 / 0.58 / 0.19 & 19.4 / 0.43 / 0.21 & 12.8 / 0.13 / 0.56 & 15.4 / 0.20 / 0.44 & 12.3 / 0.11 / 0.52 & 16.3 / 0.32 / 0.37 \\ 
        TiNeuVox~\cite{TiNeuVox} & 19.7 / 0.60 / 0.26 & 21.9 / 0.68 / 0.16 & 22.9 / 0.63 / 0.19 & 16.2 / 0.34 / 0.37 & 18.1 / 0.41 / 0.29 & 12.6 / 0.14 / 0.46 & 18.6 / 0.47 / 0.29 \\
        NSFF~\cite{li2021neural} & \textcolor{blue}{\underline{23.9}} / 0.80 / 0.15 &	28.8 / 0.88 / 0.13 & 25.4 / 0.76 / 0.17 &	21.5 / 0.69 / 0.22 & 23.8 / 0.73 / 0.23 &	20.8 / 0.70 / 0.22	& 24.1 / 0.76 / 0.18 \\
        DVS~\cite{gao2021dynamic}  & 23.4 / 0.83 / \textcolor{blue}{\underline{0.10}} & \textbf{\textcolor{red}{31.9}} / \textcolor{blue}{\underline{0.94}} / \textbf{\textcolor{red}{0.04}} & 27.9 / 0.86 / 0.09 & \textcolor{blue}{\underline{21.6}} / 0.75 / \textbf{\textcolor{red}{0.11}} & \textbf{\textcolor{red}{26.6}} /  \textcolor{blue}{\underline{0.85}} / \textbf{\textcolor{red}{0.05}} & \textcolor{blue}{\underline{23.7}} / 0.85 / \textcolor{blue}{\underline{0.08}}	& \textbf{\textcolor{red}{25.9}} / 0.85 / \textcolor{blue}{\underline{0.08}}  \\
        RoDynRF~\cite{liu2023robust} & \textbf{\textcolor{red}{24.3}} / \textcolor{blue}{\underline{0.84}} / \textbf{\textcolor{red}{0.08}} & 27.5 / 0.93 / \textcolor{blue}{\underline{0.06}} &  \textcolor{blue}{\underline{28.3}} / \textcolor{blue}{\underline{0.89}} / \textbf{\textcolor{red}{0.07}} & 21.4 / \textcolor{blue}{\underline{0.76}} / \textbf{\textcolor{red}{0.11}} & 25.6 / 0.84 / \textcolor{blue}{\underline{0.06}} & \textbf{\textcolor{red}{24.3}} / \textcolor{blue}{\underline{0.89}} / \textbf{\textcolor{red}{0.05}} & 25.2 / \textcolor{blue}{\underline{0.86}} / \textbf{\textcolor{red}{0.07}}\Bstrut \\ \hline
        \textbf{Point-DynRF (Ours)} & 23.6 / \textbf{\textcolor{red}{0.90}} / 0.14 &  \textcolor{blue}{\underline{29.6}} / \textbf{\textcolor{red}{0.96}} / \textbf{\textcolor{red}{0.04}} & \textbf{\textcolor{red}{28.5}} / \textbf{\textcolor{red}{0.94}} / \textcolor{blue}{\underline{0.08}} & \textbf{\textcolor{red}{21.7}} / \textbf{\textcolor{red}{0.88}} / \textcolor{blue}{\underline{0.12}} & \textcolor{blue}{\underline{26.2}} / \textbf{\textcolor{red}{0.92}} / \textcolor{blue}{\underline{0.06}} & 22.2 / \textbf{\textcolor{red}{0.91}} / 0.09 & \textcolor{blue}{\underline{25.3}} / \textbf{\textcolor{red}{0.92}} / \textcolor{blue}{\underline{0.08}}\Tstrut\Bstrut \\ \Xhline{3\arrayrulewidth}
        \end{tabular}}
    \label{table:quan_nvidia}
    \vspace{-0.3cm}
\end{table*}

\begin{figure*}[t!]
    \centering
    \includegraphics[width=0.135\textwidth]{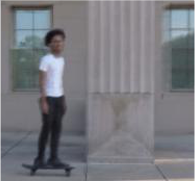}
    \includegraphics[width=0.135\textwidth]{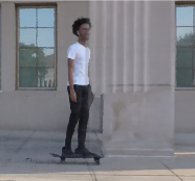}
    \includegraphics[width=0.135\textwidth]{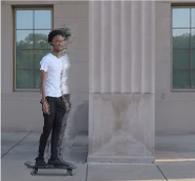}
    \includegraphics[width=0.135\textwidth]{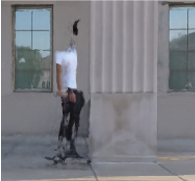}
    \includegraphics[width=0.135\textwidth]{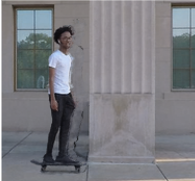}
    \includegraphics[width=0.135\textwidth]{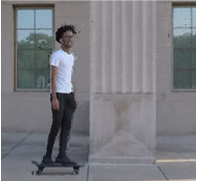}
    \includegraphics[width=0.135\textwidth]{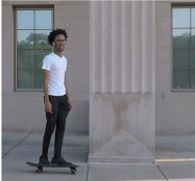}

    \includegraphics[width=0.135\textwidth]{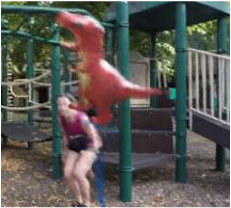}
    \includegraphics[width=0.135\textwidth]{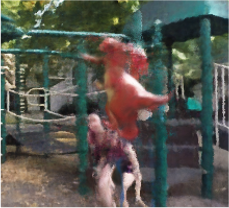}
    \includegraphics[width=0.135\textwidth]{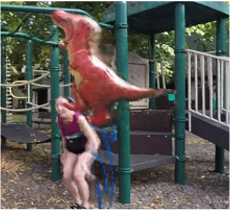}
    \includegraphics[width=0.135\textwidth]{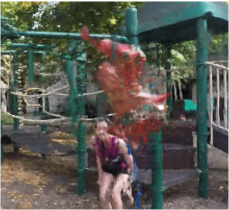}
    \includegraphics[width=0.135\textwidth]{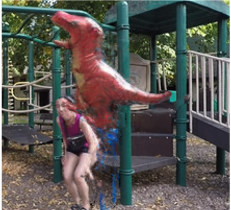}
    \includegraphics[width=0.135\textwidth]{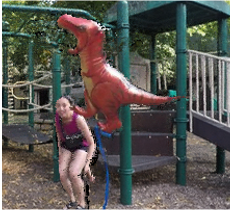}
    \includegraphics[width=0.135\textwidth]{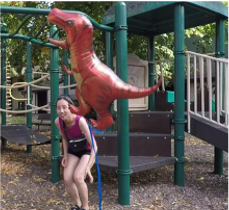}

    {\parbox{0.135\textwidth}{\centering {\small NSFF~\cite{li2021neural}}}}
    {\parbox{0.135\textwidth}{\centering {\small HyperNeRF~\cite{park2021hypernerf}}}}
    {\parbox{0.135\textwidth}{\centering {\small DVS~\cite{gao2021dynamic}}}}
    {\parbox{0.135\textwidth}{\centering {\small TiNeuVox~\cite{TiNeuVox}}}}
    {\parbox{0.135\textwidth}{\centering {\small RoDynRF~\cite{liu2023robust}}}}
    {\parbox{0.135\textwidth}{\centering {\small Ours}}}
    {\parbox{0.135\textwidth}{\centering {\small Ground Truth}}}
    
    \caption{\textbf{Comparison to baselines on NVIDIA Dynamic Scene Dataset~\cite{yoon2020novel}.}}
    \label{fig_qual}
    \vspace{-0.5cm}
\end{figure*}

\section{Experiments}

\subsection{Experimental Settings}
\paragraph{Dataset.}
We evaluate our method on the Dynamic Scene Dataset~\cite{yoon2020novel}.
We also use the same evaluation protocol in DVS~\cite{gao2021dynamic}, which evaluate the quality of the synthesized novel views through PSNR, SSIM and LPIPS metrics with ground truth images.
Note that we exclude the Umbrella sequences since COLMAP estimates inaccurate camera poses, failing to regress the scene geometry accurately.
Instead, we evaluate our method on several causally captured monocular video clips, which are more realistic videos and have a wide range of camera trajectories.
Causally captured videos provide various scene contexts that can be happened in real-world scenarios, and COLMAP accurately estimates camera poses.

\subsection{Comparison to Baselines}
We now compare our method with the state-of-the-art methods on the NVIDIA Dynamic Scene dataset~\cite{yoon2020novel}.
Table~\ref{table:quan_nvidia} shows quantitative results, and Point-DynRF demonstrates competitive performance with previous methods across most scenes.
Specifically, Point-DynRF outperforms all previous methods on the SSIM metric for all scenes.
However, due to the inaccurate camera pose estimated by COLMAP, the construction of neural points in Point-DynRF is not optimal.
As a result, the rendered position of the dynamic object by Point-DynRF may differ from the ground truth.
In the Playground scene depicted in Fig.~\ref{fig_qual}, Point-DynRF generates a visually pleasing view but the position of the object is slightly shifted behind compared to the ground-truth.
In the Skating scene, Point-DynRF generate realistic images, while flow-based methods like NSFF~\cite{li2021neural}, DVS~\cite{gao2021dynamic}, and RoDynRF~\cite{liu2023robust} produce blurry images, and deformable NeRFs such as HyperNeRF~\cite{park2021hypernerf} and TiNeuVox~\cite{TiNeuVox} struggle to represent the scene.

\begin{figure}[t]
  \centering
  \includegraphics[width=\linewidth]{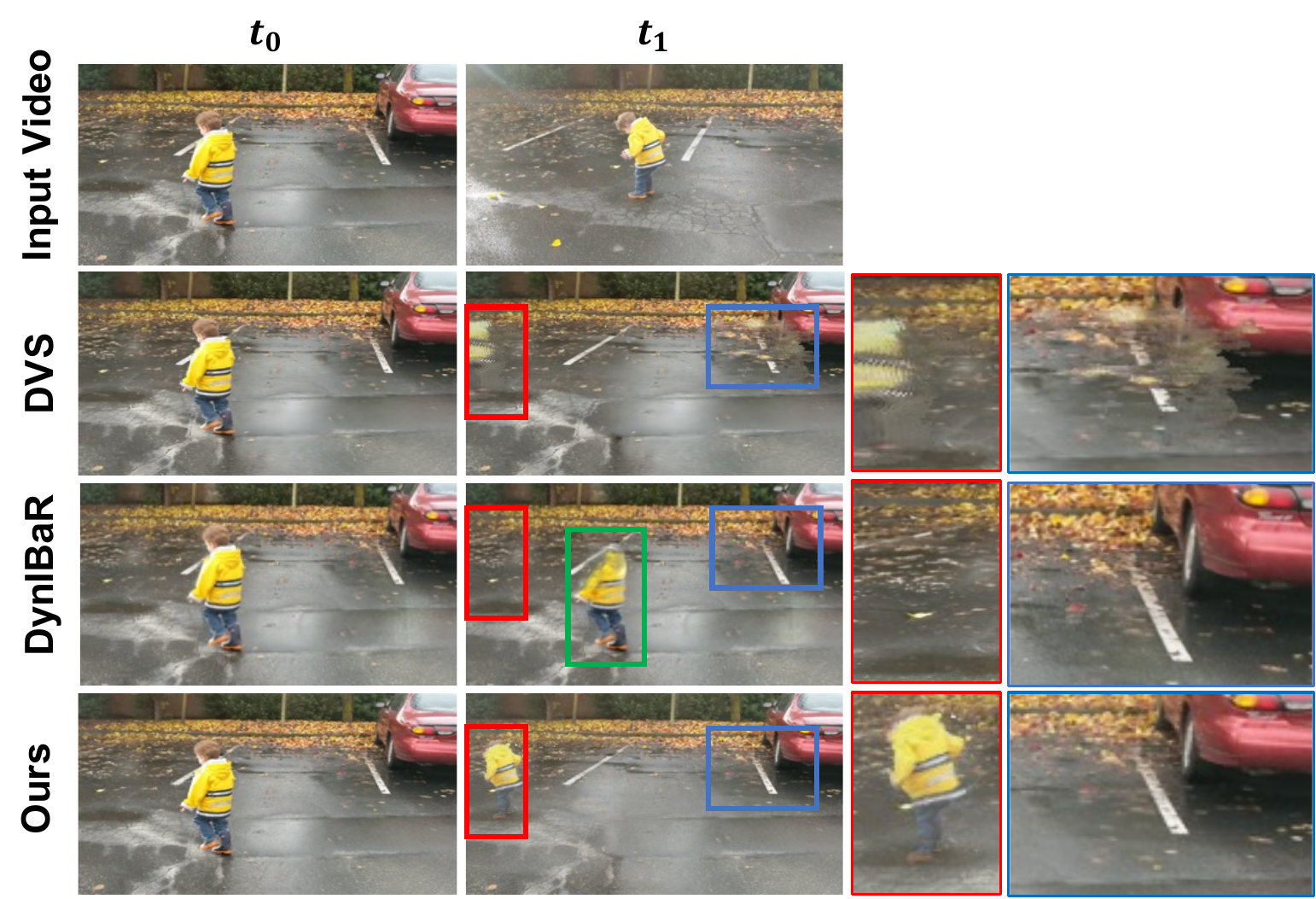}
  \vspace{-0.6cm}
  \caption{
  \textbf{Long-Term Novel View Generation.} For a fixed camera viewpoint at time $t_{0}$, the first column shows the novel view at time $t_{0}$ and the second column shows the novel view at time $t_{1}$. DynIBaR is over-fitted on the input camera trajectory (\textcolor{green}{green box})}
  \vspace{-0.6cm}
  \label{qual_result_1}
\end{figure}

\begin{figure}[t]
  \centering
  \includegraphics[width=0.95\linewidth]{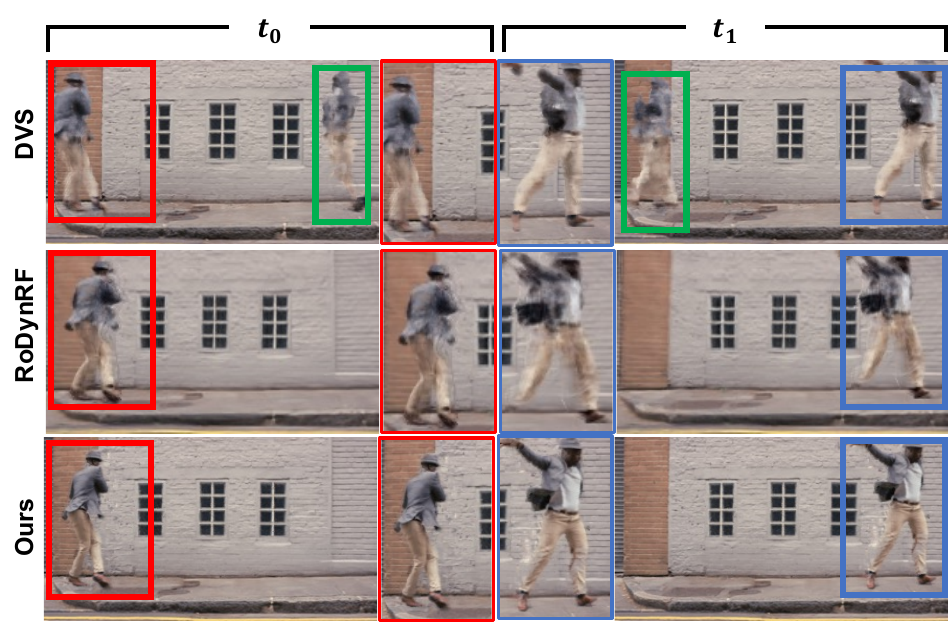}
  \vspace{-0.2cm}
  \caption{\textbf{Extremely Wide-Range Camera Trajectory.} DVS produces duplicated dynamic objects in distant spatio-temporal locations from the input camera trajectory (\textcolor{green}{green boxes}). Moreover, previous methods are quickly degenerated at the spatio-temporally distant viewpoints.}
  \label{qual_result_2}
  \vspace{-0.5cm}
\end{figure}

\subsection{Long-Term View Synthesis}
We evaluate our method and flow-based dynamic view synthesis methods DVS~\cite{gao2021dynamic}, RoDynRF~\cite{liu2023robust} and DynIBaR~\cite{li2023dynibar} on real-world scenarios with a wide-range camera trajectory.
Point-DynRF can generate realistic novel views for viewpoints far from the input camera trajectory in both space and time because it leverages global scene geometry (i.e., neural 3D points).
Figure~\ref{qual_result_1} shows the long-term view synthesis results where DVS~\cite{gao2021dynamic} has quickly degenerated for unseen viewing directions and produces artifacts in the background regions.
Moreover, DynIBaR~\cite{li2023dynibar} fails to generate a dynamic object since it is highly over-fitted on the input trajectory.
On the other hand, our Point-DynRF effectively captures both the moving object and the background.

We also observe that DVS~\cite{gao2021dynamic} infinitely duplicates the moving object and RoDynRF~\cite{liu2023robust} is also degenerated when a camera moves in an extremely single direction, as shown in Fig.~\ref{qual_result_2}.
This is due to geometric constraints focused on the input camera trajectory, which fails to represent the global scene geometry.
Notably, our Point-DynRF generates more detailed dynamic regions as well as static background regions.
These results confirm the superiority of our dynamic ray sampling and joint optimization scheme, which incorporates the entire scene geometry. 

\begin{figure}[t]
  \centering
  \includegraphics[width=0.98\linewidth]{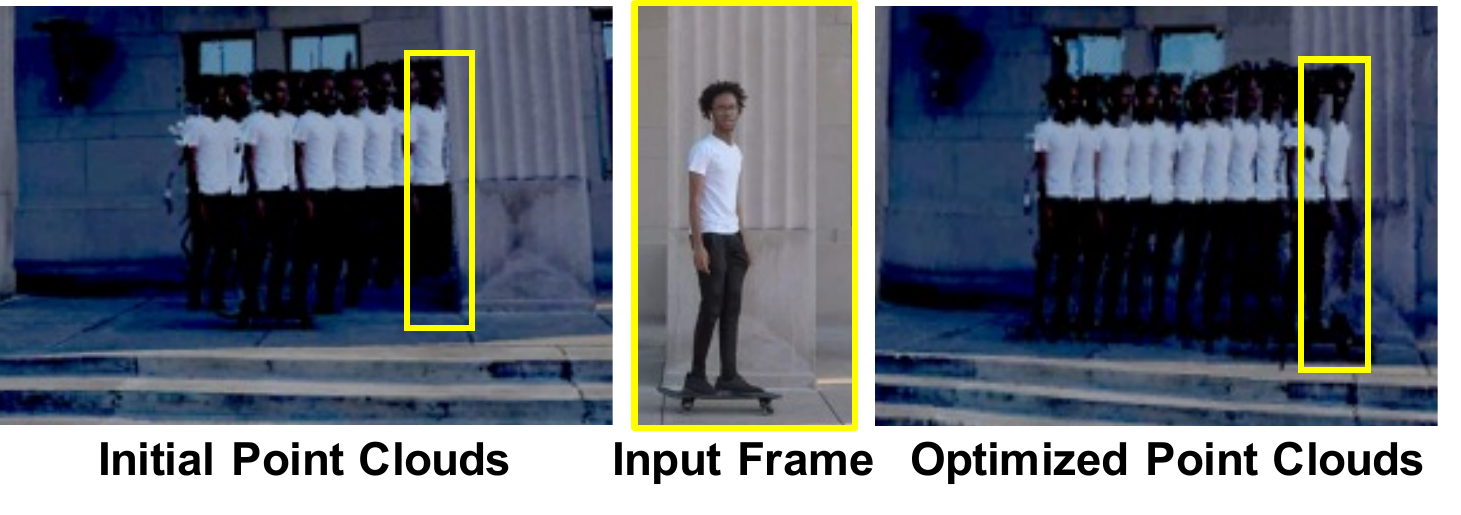}
    \vspace{-0.3cm}
  \caption{\textbf{Refinement of scale ambiguity.} The initial point cloud may not capture the complete scene geometry, but after optimization, the refined point cloud is free from scale ambiguity.}
  \label{refine_pc}
\end{figure}

\begin{figure}[t]
  \centering
  \includegraphics[width=0.24\linewidth]{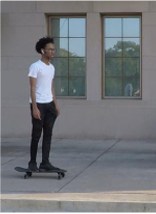}
  \includegraphics[width=0.24\linewidth]{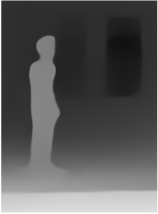}
  \includegraphics[width=0.24\linewidth]{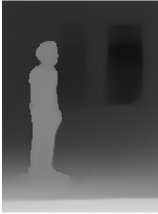}
  \includegraphics[width=0.24\linewidth]{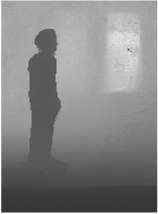}
  
  {\parbox{0.24\linewidth}{\centering {\small (a)}}}
  {\parbox{0.24\linewidth}{\centering {\small (b)}}}
  {\parbox{0.24\linewidth}{\centering {\small (c)}}}
  {\parbox{0.24\linewidth}{\centering {\small (d)}}}
  \vspace{-0.2cm}
  \caption{\textbf{Refinement of depth map.} For a input frame (a) and initialized disparity map (b), optimized disparity map (c) well represent the boundary of a dynamic object. Also, expected depth map (d) is well aligned with the disparity map.}
  \label{refine_depth}
  \vspace{-0.4cm}
\end{figure}

\subsection{Effect of Accurate Scene Geometry}
To verify the effectiveness of our proposed losses for training the scene geometry,  we visualize initialized and refined point clouds as well as a depth map on the Skating scene as shown in Fig.~\ref{refine_pc}-\ref{refine_depth}.
The results demonstrate that our joint optimization effectively regresses the scene geometry and address the scale ambiguity problem in monocular videos, resulting in a dynamic radiance field that accurately reflects this geometry.

\subsection{Training and Rendering Time}

Table~\ref{table:times} shows the training and rendering time on NVIDIA Dynamic Scene Dataset~\cite{yoon2020novel} for recent dynamic view synthesis methods.
Since Point-DynRF avoids the unnecessary ray marching for empty space, the training process converges faster, leading to a reduction in overall training time.
In the rendering process, however, searching neighbor neural points for each shading point requires additional computational costs, and the rendering time is slower than DVS~\cite{gao2021dynamic} and RoDynRF~\cite{liu2023robust}.

\begin{table}[ht]
\caption{\textbf{Comparison of Training and Rendering Time.} Methods denoted by $\dagger$ refer to reported performance in the paper.}
    \centering
    \resizebox{\linewidth}{!}{
    \begin{tabular}{l|c|c}
    \hline
    Method & Training (GPU hours) & Rendering (s/img) \\
    \hline
    HyperNeRF~\cite{park2021hypernerf} & 32 & 15 \\
    DVS~\cite{gao2021dynamic} & 36 & 8 \\
    $\text{RoDynRF}^\dagger$~\cite{liu2023robust} & 28 & 8 \\
    $\text{DynIBaR}^\dagger$~\cite{li2023dynibar} & 48 & 20 \\
    Ours & 20 & 11 \\
    \hline
    \end{tabular}
    }
   
\label{table:times}
\end{table}

\subsection{Ablation Study for Point-DynRF Design}
\begin{figure}[t]
  \centering
  \includegraphics[width=\linewidth]{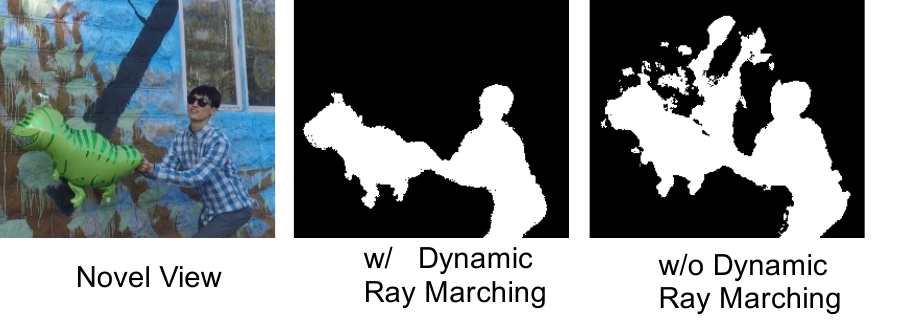}
  \vspace{-0.6cm}
  \caption{
  \textbf{Qualitative Ablation of Dynamic Ray Marching.} Without the dynamic ray marching, dynamic points at other times interfere with dynamic radiance fields at the rendering time.}
  \label{dynamicsness_map}
\end{figure}

We conduct an ablation study for each component of Point-DynRF as shown in Table~\ref{table:ablation}.
The results show the quantitative results, and we verify all components contribute to the design of our Point-DynRF.
Especially from the results on $L_{flow}$ and $L_{depth}$, we confirm that the accuracy of the neural points has a direct impact on the performance of dynamic radiance fields.
Also, the results on $\mathbb{P}_{t}$ confirm that our dynamic ray marching scheme significantly improves the performance.
Dynamic ray marching ensures that the dynamicsness map for the novel view is well matched to the actual scene, as shown in Fig.~\ref{dynamicsness_map}.

\begin{table}[t!]
 \caption{\textbf{Ablation Study of our proposed losses.} We report the PSNR, SSIM and LPIPS on the average of NVIDIA Dynamic Scene Dataset~\cite{yoon2020novel}.}
 
    \centering
    \resizebox{\linewidth}{!}{
    \begin{tabular}{l|ccc}
\Xhline{3\arrayrulewidth}
  & PSNR ($\uparrow$)& SSIM ($\uparrow$) & LPIPS ($\downarrow$)\\ \hline

Ours w/o $\mathbb{P}_{t}$      & 23.62 \textbf{\textcolor{red}{(-1.68)}} & 0.755 \textbf{\textcolor{red}{(-0.161)}} & 0.148 \textbf{\textcolor{red}{(+0.067)}} \\
Ours w/o $L^{s}_{rec}$       & 24.38 \textbf{\textcolor{red}{(-0.92)}} & 0.843 \textbf{\textcolor{red}{(-0.073)}} & 0.121 \textbf{\textcolor{red}{(+0.040)}} \\
Ours w/o $L^{d}_{rec}$      & 25.08 \textbf{\textcolor{red}{(-0.22)}} & 0.901 \textbf{\textcolor{red}{(-0.015)}} & 0.097 \textbf{\textcolor{red}{(+0.016)}} \\
Ours w/o $L_{flow}$       & 24.13 \textbf{\textcolor{red}{(-1.07)}} & 0.872 \textbf{\textcolor{red}{(-0.042)}} & 0.099 \textbf{\textcolor{red}{(+0.018)}} \\
Ours w/o $L_{depth}$      & 24.45 \textbf{\textcolor{red}{(-0.85)}} & 0.856 \textbf{\textcolor{red}{(-0.070)}} & 0.100 \textbf{\textcolor{red}{(+0.019)}} \\
Ours w/o $L_{mask}$      & 24.66 \textbf{\textcolor{red}{(-0.64)}} & 0.884 \textbf{\textcolor{red}{(-0.032)}} & 0.111 \textbf{\textcolor{red}{(+0.030)}} \\
Ours                     & \textbf{25.30}  & \textbf{0.916} & \textbf{0.081} \\ \Xhline{3\arrayrulewidth}
\end{tabular}
}
 \label{table:ablation}
 \vspace{-0.2cm}
\end{table}

\section{Conclusion}
We propose a novel framework called point-based dynamic radiance fields for long-term dynamic view synthesis from monocular videos.
In our approach, we employ neural point clouds to encode geometric information and dynamic radiance fields to handle the volume rendering process.
Our framework, Point-DynRF, optimizes the neural point clouds and dynamic radiance fields jointly, leveraging direct regression from neural 3D points.
This allows us to effectively utilize the global scene geometry, which sets our method apart from previous approaches relying on correspondences between neighboring frames, limiting their ability to incorporate the overall scene geometry.
We believe that our work contributes significantly to the field of dynamic view synthesis, enabling realistic rendering in various real-world scenarios.

\appendix

\section{Overview}
In this supplementary material, we further demonstrate our experimental setup and provide additional results that the scene geometry is well regressed.
First, we explain the total loss formulation in our training process in Sec.~\ref{losses}.
Then, we describe implementation details with image near-far bound determination by neural points in Sec.~\ref{imple} and provide additional results for dynamicsness map of novel views in Sec.~\ref{results}.
Finally, we demonstrate failure cases in Sec.~\ref{failure}.

\section{Losses}
\label{losses}

Our optimization process involves utilizing the loss functions $L_{rec}$, $L_{geo}$, $L_{depth}$, and $L_{mask}$. 
These loss functions are either modifications of those used in DVS~\cite{gao2021dynamic} or newly introduced in this paper.
To train Point-DynRF more stable, we also incorporate with a depth order loss $L_{order}$ introduced in DVS~\cite{gao2021dynamic} and a sparsity loss $L_{sparse}$ introduced in Point-NeRF~\cite{xu2022point}.

\paragraph{Depth Order Loss}
While the depth adjust loss helps optimize the overall scene geometry, there are inherent challenges in accurately determining the distance between dynamic objects and the background.
Therefore, we use depth order loss $L_{order}$ to allow the dynamic radiance fields to be regularized via a frame-by-frame depth map.
Since regularizing the dynamic radiance fields with per-frame depth maps has scale and shift ambiguities as mentioned earlier, we leverage the volume rendering process of Dynamic NeRF to propose $L_{order}$ as:

\begin{equation}
L_{order} = \sum^{N}_{i=1}\sum_{uv}\norm{\textbf{\~{D}}(\textbf{r}^{i}_{uv}, i, \mathbb{P}_{i}) - \textbf{\~{D}}^{d}(\textbf{r}^{i}_{uv}, i, \mathbb{P}_{i, d})}^{2}_{2}.
\end{equation}

\paragraph{Sparsity Loss}
Following the point-based representation, we apply a sparsity loss $L_{sparse}$ on the point-wise rigidness to enforce it to be close to zero or one as:

\begin{equation}
L_{sparse} = \sum_{i}(\log(\gamma_{i}) + \log(1 - \gamma_{i})).
\end{equation}

\paragraph{Total Training Loss Formulation}
We formulate a reconstruction loss $L_{rec}$, a scene geometry loss $L_{geo}$, a depth adjust loss $L_{depth}$, a depth order loss $L_{order}$, a mask adjust loss $L_{mask}$ and a sparsity loss $L_{sparse}$, to train our Point-DynRF and neural points.
Specifically, we define $\lambda^{full}_{rec} = 3$, $\lambda^{s}_{rec} = 1$, $\lambda^{d}_{rec} = 1$ for the reconstruction loss.
For the scene geometry loss, we define $\lambda_{flow} = 0.1$, $\lambda^{s}_{miss} = 1$, $\lambda^{d}_{miss} = 1$.
Finally, we define $\lambda_{depth} = 0.1$, $\lambda_{order} = 0.1$, $\lambda_{mask} = 0.1$, and $\lambda_{sparse} = 0.0002$ to formulate the final loss as:

\begin{align*}
L_{total} &= L_{rec} + L_{geo} + \lambda_{depth}L_{depth} + \lambda_{order}L_{order} + \\ 
          & \qquad \lambda_{mask}L_{mask} + \lambda_{sparse}L_{order}.
\end{align*}

\section{Implementation Details.}
\label{imple}
We randomly sampled $1024$ rays in a batch, and each ray was assigned up to $32$ sampling points.
We used COLMAP to estimate the camera poses and resized all images into a resolution of $480 \times 272$.
Also, we initialized our scale and shift parameters by using near and far bounds from COLMAP.
We trained Point-DynRF for 250$k$ iterations, and training takes about $20$ hours on a single NVIDIA Geforce RTX $3090$ GPU.

\begin{figure}[t]
  \centering
  \includegraphics[width=\linewidth]{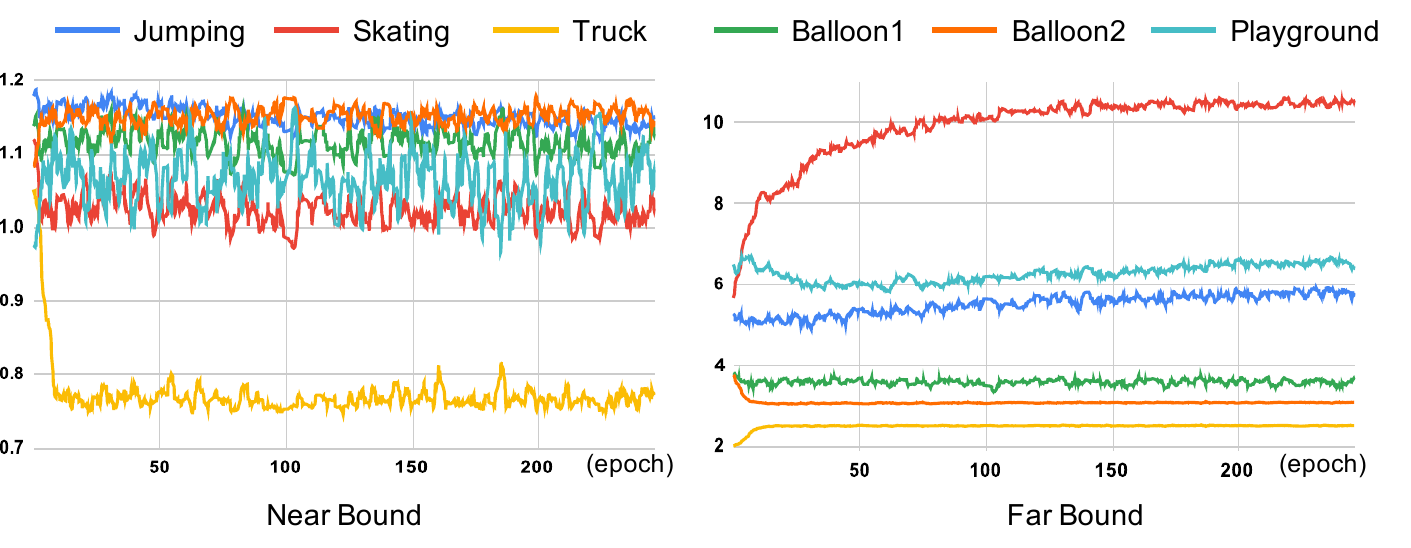}
  \vspace{-0.6cm}
  \caption{
  \textbf{Image Near-Far Bound Determination.}}
  \vspace{-0.4cm}
  \label{near-far}
\end{figure}

\begin{figure*}[t!]
    \centering
    \includegraphics[width=0.16\textwidth]{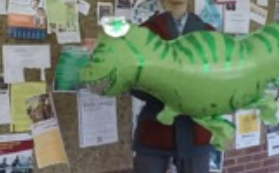}
    \includegraphics[width=0.16\textwidth]{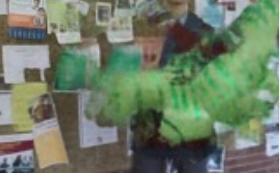}
    \includegraphics[width=0.16\textwidth]{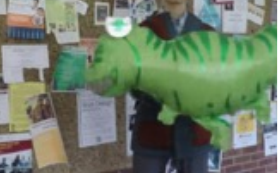}
    \includegraphics[width=0.16\textwidth]{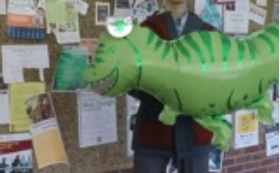}
    \includegraphics[width=0.16\textwidth]{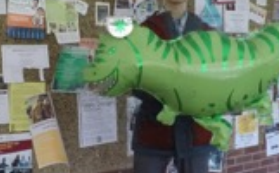}
    \includegraphics[width=0.16\textwidth]{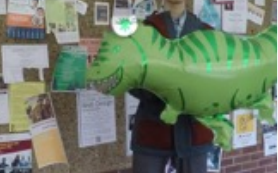}
    
    \includegraphics[width=0.16\textwidth]{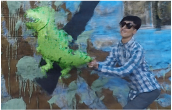}
    \includegraphics[width=0.16\textwidth]{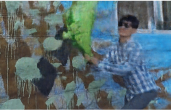}
    \includegraphics[width=0.16\textwidth]{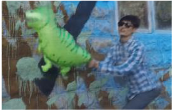}
    \includegraphics[width=0.16\textwidth]{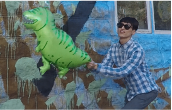}
    \includegraphics[width=0.16\textwidth]{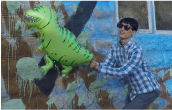}
    \includegraphics[width=0.16\textwidth]{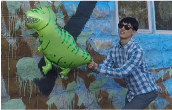}

    \includegraphics[width=0.16\textwidth]{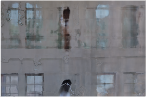}
    \includegraphics[width=0.16\textwidth]{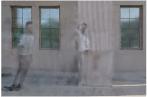}
    \includegraphics[width=0.16\textwidth]{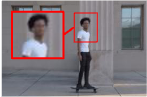}
    \includegraphics[width=0.16\textwidth]{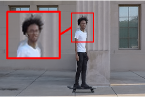}
    \includegraphics[width=0.16\textwidth]{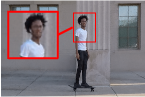}
    \includegraphics[width=0.16\textwidth]{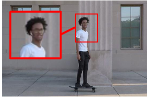}

    {\parbox{0.16\textwidth}{\centering {\small NeRF $+$ time~\cite{mildenhall2021nerf}}}}
    {\parbox{0.16\textwidth}{\centering {\small NR-NeRF~\cite{tretschk2021non}}}}
    {\parbox{0.16\textwidth}{\centering {\small NSFF~\cite{li2021neural}}}}
    {\parbox{0.16\textwidth}{\centering {\small DVS~\cite{gao2021dynamic}}}}
    {\parbox{0.16\textwidth}{\centering {\small Ours}}}
    {\parbox{0.16\textwidth}{\centering {\small Ground Truth}}}
    
    \caption{\textbf{Comparison to baselines on NVIDIA Dynamic Scene Dataset~\cite{yoon2020novel}.}}
    \vspace{-0.4cm}
    \label{fig_quall}
\end{figure*}

\paragraph{Near-Far Boundary Determination}
As our Point-DynRF is built on Point-NeRF~\cite{xu2022point} representation, dynamic radiance fields are regressed in 3D world coordinates, not in NDC space used by previous methods.
Moreover, we need to render the far background as well, so we set the image near-far bound dynamically associated with the neural points.
Specifically, we set the image near boundary to be the depth for the nearest neural point multiplied by $0.9$, and the image far boundary to be the depth for the farthest neural point multiplied by $1.1$.
Figure~\ref{near-far} shows the convergence of the image near-far boundary of the scenes in the Dynamic Scene Dataset~\cite{yoon2020novel} during training.
This result confirms that the scene geometry is stably trained and refined the initialized scene geometry well.

\begin{figure}[t]
  \centering
  \includegraphics[width=\linewidth]{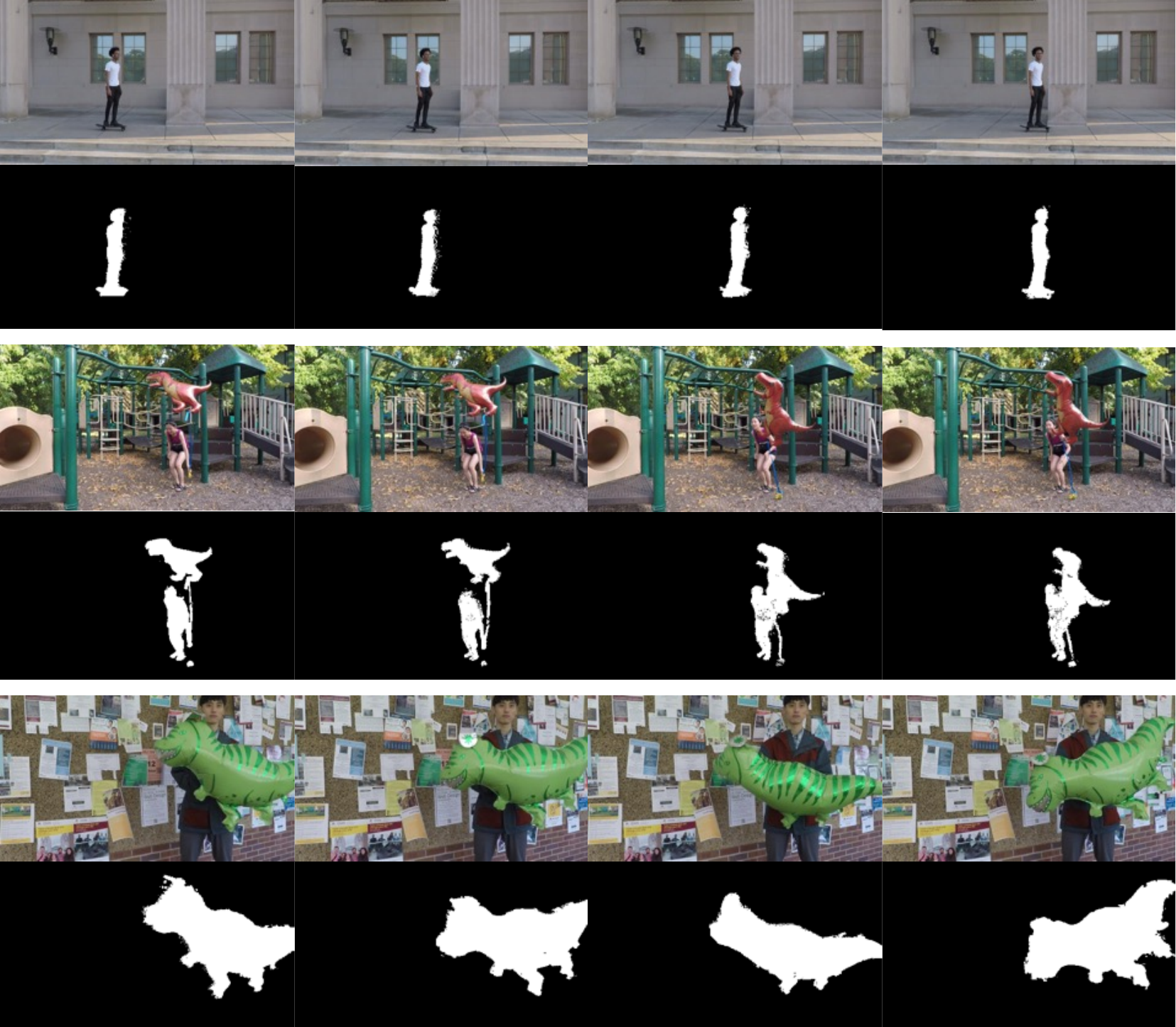}
  \vspace{-0.6cm}
  \caption{
  \textbf{Dynamicsness Maps for novel views.}}
  \vspace{-0.6cm}
  \label{dynamic_results}
\end{figure}

\section{Additional Results}
\label{results}

\paragraph{Additional Qualitative Results.}
We further provide additional qualitative results on Dynamic Scene Dataset~\cite{yoon2020novel}
. 
Point-DynRF generates more realistic images compared to previous methods, and the human face in the third row of Fig.~\ref{fig_quall} confirms that Point-DynRF produces much sharper images, while other methods either fail to synthesize or produce blurry images.
We also provide a video result of a causally captured monocular video that our Point-DynRF generates realistic images while the state-of-the-art method DVS~\cite{gao2021dynamic} suffers from duplicated dynamic objects when rendering from a fixed viewpoint.

Our foreground masks $(M_{1}, \dots, M_{N})$ are also optimized during the training, so we provide dynamicsness maps for novel views, as shown in Fig~\ref{dynamic_results}.
For each novel view, our Point-DynRF can render blending weights by using the volume rendering process.
These dynamicsness maps for novel views confirm that our Point-DynRF well represents dynamic regions in the scene, and we can see that the static representation in the center of the person in the Playground Sequence is due to the fact that all the sequences in the input video for that region are learned as dynamic regions and represented as background by the miss ray marching scheme.

\section{Failure Cases}
\label{failure}

\begin{figure}[h!]
  \centering
  \includegraphics[width=0.9\linewidth]{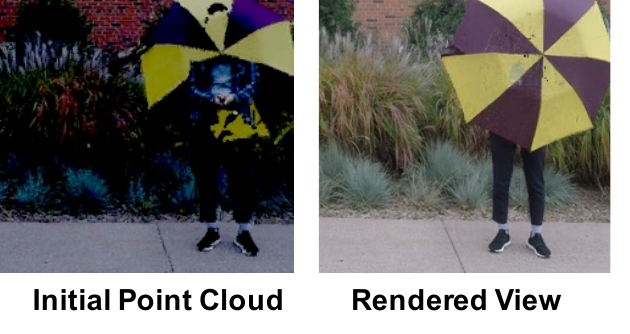}
  \vspace{-0.4cm}
  \caption{
  \textbf{Failure Case.}}
  \vspace{-0.4cm}
  \label{fail_umb}
\end{figure}

While Point-DynRF optimizes well the ambiguous initial geometry and foreground masks, it fails to represent the scene if the neural point clouds are unnaturally initialized.
A combination of inaccurate camera pose, depth map, and foreground masks sometimes unnaturally initialize neural point clouds where background points are closer to the camera than dynamic points as shown in Fig.~\ref{fail_umb}.
In this failure case, Point-DynRF falls short of distinguishing background points in front of the dynamic objects even addressing the scale ambiguity, and novel views also contain artifacts on these background points.

{\small
\bibliographystyle{ieee_fullname}
\bibliography{egbib}
}

\end{document}